\documentclass[conference]{IEEEtran}

\usepackage{times}
\usepackage{epsfig}
\usepackage{graphicx}
\usepackage{amsmath}
\usepackage{amssymb}
\usepackage{algpseudocode}
\usepackage{algorithm}
\usepackage{array}
\usepackage{multirow}
\usepackage{tabularx}

\usepackage{setspace}

\usepackage{booktabs}

\usepackage{url}
\usepackage{breakurl}

\usepackage{fancyhdr}

\newcolumntype{Y}{>{\raggedleft\arraybackslash}X}

\graphicspath{ {images/} }
\usepackage{pdfpages}
\pdfoutput=1

\begin{document}

\title{CP-decomposition with Tensor Power Method for Convolutional Neural Networks Compression}
\author{\IEEEauthorblockN{Marcella Astrid}
	\IEEEauthorblockA{University of Science and Technology\\
		Daejeon, South Korea\\
		Email: marcella.astrid@ust.ac.kr}
	\and
	\IEEEauthorblockN{Seung-Ik Lee}
	\IEEEauthorblockA{Electronics and Telecommunications Research Institute\\
		University of Science and Technology\\
		Daejeon, South Korea\\
		Email: the\_silee@etri.re.kr}}
\maketitle
\thispagestyle{fancy}
\fancyhf{}
\lhead{Accepted as a conference paper at BigComp 2017}

\begin{abstract}
   Convolutional Neural Networks (CNNs) has shown a great success in many areas including complex image classification tasks. However, they need a lot of memory and computational cost, which hinders them from running in relatively low-end smart devices such as smart phones. We propose a CNN compression method based on CP-decomposition and Tensor Power Method. We also propose an iterative fine tuning, with which we fine-tune the whole network after decomposing each layer, but before decomposing the next layer. Significant reduction in memory and computation cost is achieved compared to state-of-the-art previous work with no more accuracy loss.
\end{abstract}

\newcommand\mycomment[1]{\textcolor{red}{#1}}
\newcommand{\Xa}{\mathcal{X}}
\newcommand{\Ta}{\mathcal{T}}
\newcommand{\Ya}{\mathcal{Y}}
\newcommand{\Ka}{\mathcal{K}}
\newcommand{\Wa}{\mathcal{W}}
\newcommand{\Ua}{\mathcal{U}}
\newcommand{\Za}{\mathcal{Z}}
\newcommand{\newln}{\\&\quad\quad{}}

\section{Introduction} \label{sec:Introduction}
Convolutional neural networks (CNNs) have shown notable results in image recognition: VGG \cite{simonyan2014very} and GoogleNet \cite{szegedy2015going} achieved around 90\% accuracy for top-5 classification in ImageNet2012 dataset; AlexNet \cite{krizhevsky2012imagenet} also achieved around 80\% top-5 accuracy with the same dataset.

On the other hand, there is an emerging need for embedding or executing CNNs on smart devices such as mobile phones, robots, and embedded devices, in order to give them more intelligence. But the barrier here is that CNNs will require high amounts of memory and computational resources, which are unfortunately hard to be met by small-sized smart devices.

In order to tackle this problem, several approaches recently have been proposed based on tensor decomposition, including Tucker decomposition \cite{kim2015compression} and Canonical Polyadic (CP) decomposition \cite{denton2014exploiting,lebedev2014speeding}. Kim et al. \cite{kim2015compression} successfully decompose all the layers by using Tucker decomposition. However, Tucker decomposition does not seem to compress as much as CP-decomposition due to the core tensors. Moreover, CP-decomposition \cite{denton2014exploiting,lebedev2014speeding} has not been successful in compressing the whole convolution layers of a CNN because of the CP instability issue \cite{lebedev2014speeding}.

In this work, we further investigate low-rank CP-decomposition to compress the whole convolution layers of a CNN. Our method, called CP-TPM, is based on low-rank CP-decomposition with Tensor Power Method (TPM) for efficient optimization. We also propose iterative fine-tuning to overcome the CP-decomposition instability. We expect the followings with our method:
\begin{itemize}
	\item \textbf{The whole convolution layer decomposition with CP:} To the best of our knowledge, the whole convolution layer decomposition has not been successful because of CP-decomposition's instability \cite{lebedev2014speeding}. We expect that CP-TPM can decompose the whole convolution layers in contrast to previous CP-decomposition approaches \cite{denton2014exploiting,lebedev2014speeding}.
	
	\item \textbf{Overcoming CP-decomposition instability by iterative fine-tuning:} The instability of CP-decomposition, in our view, is the cause of ill-training, such that the loss does not decrease or even amplified when all of the layers are decomposed by CP and fine-tuned once at the final stage. We believe that CP-decomposition on the whole convolution layers without fine-tuning in between causes the loss to be accumulated and become unrecoverable by the once-and-for-all fine-tuning after decomposition. We empirically prove in the experiment that this instability can be overcome by iterative fine-tuning. 
\end{itemize}

Our CP-TPM achieves much reduction in memory and computational cost without a very small accuracy drop.

\section{Method} \label{sec:Method}

Our approach has two main steps: decomposition and fine-tuning. We apply the two steps layer-by-layer until all the layers of a CNN are decomposed and fine-tuned. All the layers except the fully connected layers are decomposed by CP decomposition, while the fully connected layers are decomposed by SVD. After each decomposition, the whole network is fine-tuned by back propagation.

Before starting the details, these are notations that we use in this paper. Tensors will be notated in calligraphy font capital letters, e.g. $\Xa$. Matrices will be notated in bold capital letters, e.g. $\mathbf{X}$. Vectors will be notated in bold small letters, e.g. $\mathbf{x}$. Scalars will be notated in regular font small letters, e.g. $x$. Regular font capital letters, e.g. $X$, will be used for dimension size.

\subsection{Kernel Tensor Decomposition} \label{subsec:KernelTensorDecomposition}

CP decomposes a tensor as a linear combination of rank one tensors (\ref{eq:CPDecomposition}). The number of components is the tensor rank $R$. Each component is an outer product of $n$ vectors, where $n$ corresponds to the number of ways of the target tensor. Rank $R$ determines the amount of weight reduction, i.e., the smaller the R is, the more reduced the weights in a convolution layer. 

\begin{equation} \label{eq:CPDecomposition}
\Xa = \sum_{r=1}^{R} \mathbf{a_r \otimes b_r \otimes c_r}
\end{equation}

\textbf{Convolution kernel tensor:} In general, convolution layers in CNNs map a 3-way input tensor $\Xa$ of size $S \times W \times H$ into a 3-way output tensor $\Ya$ of size $S \times W' \times H'$ using a 4-way kernel tensor $\Ka$ of size $T \times S \times D \times D$ with $T$ corresponding to different output channels, $S$ corresponding to different input channels, and the last two dimensions corresponding to the spatial dimension (for simplicity, we assume square shaped kernels and odd $D$) 

\begin{equation} \label{eq:Convolution}
\Ya_{t,w',h'} = \sum_{s=1}^{S} \sum_{j=1}^{D} \sum_{i=1}^{D} \Ka_{t,s,j,i} \Xa_{s,w_j,h_i} 
\end{equation}

\begin{center}
	$ w_j = (w'-1) \bigtriangleup + j - p $ and $h_i = (h'-1) \bigtriangleup + i - p $,
\end{center}

\noindent
where $\bigtriangleup$ is stride and $p$ is zero-padding size. 

\textbf{CP decomposition:} Now the problem is to approximate the kernel tensor $\Ka$ with rank-$R$ CP-decomposition. This can be represented as in (\ref{eq:Kernel}). Spatial dimensions are not decomposed as they are relatively small (e.g., $3 \times 3$ or $5 \times 5$).

\begin{equation} \label{eq:Kernel}
\Ka_{t,s,j,i} = \sum_{r=1}^{R} \mathbf{U}_{r,s}^{(1)} \Ua_{r,j,i}^{(2)} \mathbf{U}_{t,r}^{(3)}
\end{equation}

\noindent
where $\mathbf{U}_{r,s}^{(1)}$, $\Ua_{r,j,i}^{(2)}$, and $\mathbf{U}_{t,r}^{(3)}$ are the three components of sizes $R \times S$, $R \times D \times D$, and $T \times R$, respectively.

Substituting (\ref{eq:Kernel}) into (\ref{eq:Convolution}) and performing simple manipulations gives (\ref{eq:DecomposedConvolution}) for the approximate evaluation of the convolution (\ref{eq:Convolution}) from the input tensor $\Xa$ into the output tensor $\Ya$.

\noindent

\begin{equation} \label{eq:DecomposedConvolution}
\Ya_{t,w',h'} = \sum_{r=1}^{R} \mathbf{U}_{t,r}^{(3)} (\sum_{j=1}^{D} \sum_{i=1}^{D} \Ua_{r,j,i}^{(2)} (\sum_{s=1}^{S} \mathbf{U}_{r,s}^{(1)} \Xa_{s,w_j,h_i}))
\end{equation}

\begin{figure}[h]
	\centering
	\includegraphics[width=\linewidth]{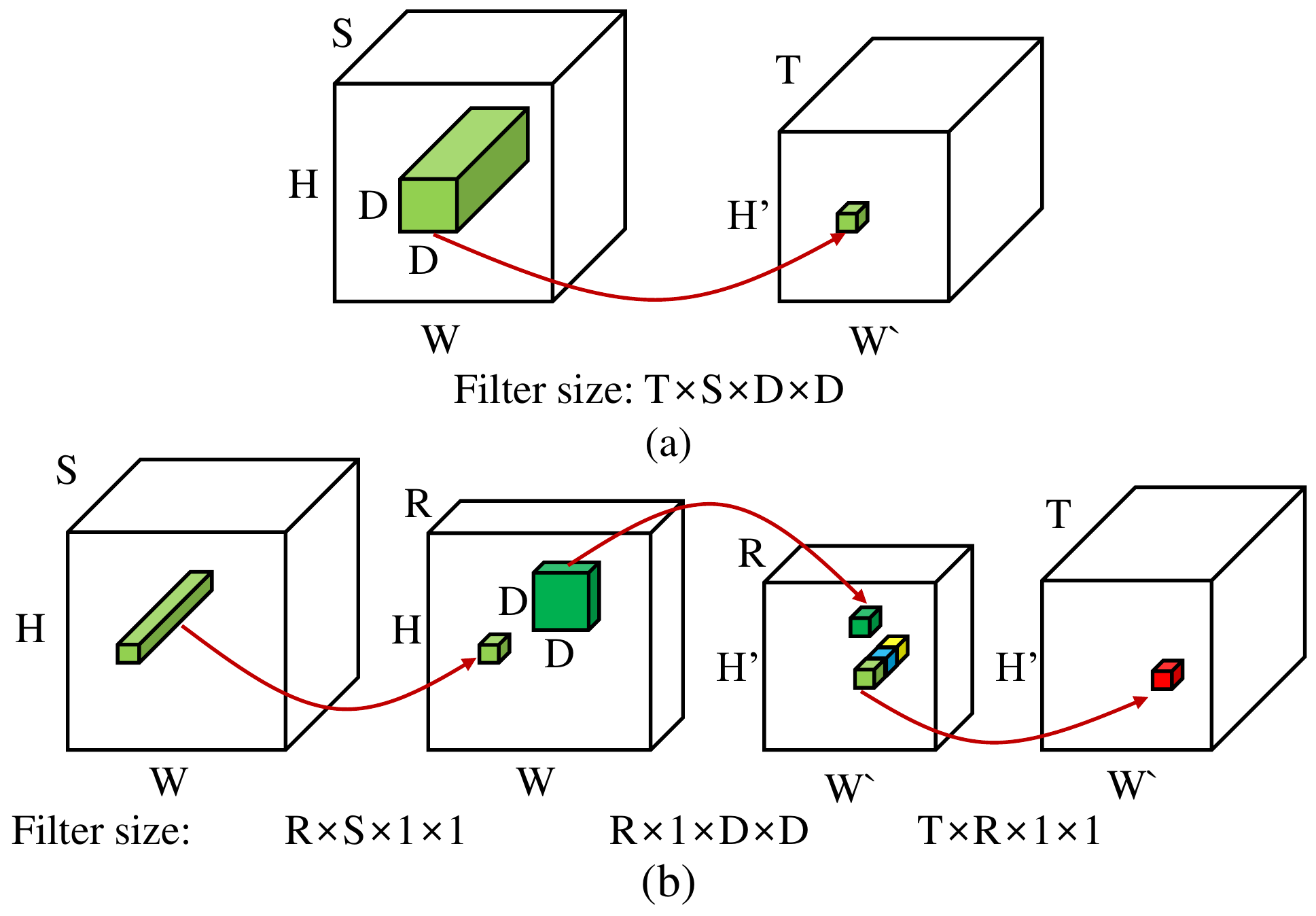}
	\caption{Convolution layer. \textbf{(a)} Original convolution layer. \textbf{(b)} CP decomposed convolution layer in this paper.}
	\label{fig:ConvGroup1}
\end{figure}

Equation (\ref{eq:DecomposedConvolution}) tells us that the output tensor $\Ya$ is computed by a sequence of three separate convolution operations from the input tensor $\Xa$ with smaller kernels (Fig. \ref{fig:ConvGroup1}(b)):

\begin{equation} \label{eq:Convolution1}
\Za_{r,w,h} = \sum_{s=1}^{S} \mathbf{U}_{r,s}^{(1)} \Xa_{s,w,h}
\end{equation}

\begin{equation} \label{eq:Convolution2}
\Za_{r,w',h'}^{'} = \sum_{j=1}^{D} \sum_{i=1}^{D} \Ua_{r,j,i}^{(2)} \Za_{t,w_j,h_i}
\end{equation}

\begin{equation} \label{eq:Convolution3}
\Ya_{t,w',h'} = \sum_{r=1}^{R} \mathbf{U}_{t,r}^{(3)} \Za_{r,w',h'}^{'}
\end{equation}

\noindent
where $\Za_{r,w,h}$ and $\Za_{r,w',h'}^{'}$ are intermediate tensors of sizes $R \times W \times H$ and $R \times W' \times H'$, respectively.

\textbf{Fully Connected Layers (FC):} Fully connected layer calculation has the form in (\ref{eq:FC}):

\begin{equation} \label{eq:FC}
\mathbf{y}^T = \mathbf{x}^T \mathbf{W}
\end{equation}

\noindent
where $\mathbf{y}^T$ is the transpose vector of the output of size $M$, $\mathbf{x}^T$ is the transpose vector of the input of size $N$, and $\mathbf{W}$ is a weight matrix of size $M \times N$. 

\textbf{Decomposition of FC:} As the weights are in a matrix form, we apply Singular Value Decomposition (SVD) to decompose the weight matrix as in (\ref{eq:SVD}):

\begin{equation} \label{eq:SVD} 
\mathbf{W} = \mathbf{U} \mathbf{D} \mathbf{V}^{T} = (\mathbf{U} \mathbf{D}) \mathbf{V}^{T}
\end{equation}

\noindent
where $\mathbf{U}$ and $\mathbf{V}^T$ are left and right singular matrices of sizes $M \times R$ and $R \times N$, respectively, and $D$ is a $R \times R$ diagonal singular-value matrix. 

Substituting (\ref{eq:SVD}) into (\ref{eq:FC}) and performing grouping gives (\ref{eq:SVDFC}) for approximate evaluation of a fully connected layer with smaller weight matrices.

\begin{equation} \label{eq:SVDFC}
\mathbf{y}^T = (\mathbf{x}^T(\mathbf{UD})) \mathbf{V}^T
\end{equation}

Therefore, one fully connected layer can be represented as two fully connected layers as in (\ref{eq:SVDFC1}) and (\ref{eq:SVDFC2}):

\begin{equation} \label{eq:SVDFC1}
\mathbf{z}^T = \mathbf{x}^T(\mathbf{UD})
\end{equation}
\begin{equation} \label{eq:SVDFC2}
\mathbf{y}^T = \mathbf{z}^T \mathbf{V}^T
\end{equation}

\noindent
where $\mathbf{z}^T$ is an intermediate layer of size $R$.

\subsection{Complexity Analysis} \textbf{\label{subsec:ComplexityAnalysis}}

The initial convolution operation in (\ref{eq:Convolution}) requires $TSD^2$ parameters and $TSD^2W'H'$ multiplication operation. With CP decomposition, the compression ratio $E$ and speed-up ratio $C$ are given by:

\begin{equation} \label{eq:Compression1}
E = \frac{TSD^2}{RS + RD^2 + TR}
\end{equation}
\begin{equation} \label{eq:SpeedUp1}
C = \frac{TSD^2W'H'}{RSWH + RD^2W'H' + TRW'H'}
\end{equation}

The initial operations in FC of (\ref{eq:FC}) is defined by $MN$ parameters and requires the same number of \textit{multiplication-addition} operations. Therefore, the compression ratio $E$ and speed-up ration $C$ are the same and given by:

\begin{equation} \label{eq:SpeedUp3}
E = C = \frac{MN}{MR + RN}
\end{equation} 

\subsection{Rank Selection} \label{subsec:RankSelection}

Ranks play a key role in CP decomposition. If the rank is too high, compression would not be maximized, and if it is too low, the accuracy would drop too much to be recovered by fine-tuning. However, there is no straight algorithm to find the optimal tensor rank \cite{kolda2009tensor}. In fact, determining the rank is NP-hard \cite{hillar2013most}. 

Thus, we apply a primitive principle in determining the rank: the higher the accuracy loss caused by a layer, the higher rank the layer needs. Rank proportion is the proportion of rank of a layer among other layers. In order to figure out how sensitive a layer is to decomposition, we perform kind of prior decomposition of each layer with a very low, but constant rank (e.g., 5), and then fine-tune the whole network (one epoch).

\subsection{Computation of Tensor Decomposition} \label{subsec:ComputationOfTensorDecomposition}
In general, tensor decomposition is an optimization problem, i.e., minimizing the difference between the decomposed tensor and the target tensor. We employ Tensor Power Method (TPM) \cite{allen2012sparse}. TPM is known to explain the same variance with less rank compared to ALS \cite{allen2012sparse} because the rank-1 tensors found in the early steps of the process explains most of the variances in the target tensor.  

TPM approximates a target tensor $\Wa$ by adding rank-1 tensors iteratively. First, TPM finds a rank-1 tensor, $\Wa_{decomposed}$, to approximate $\Wa$ by minimizing $||\Wa - \Wa_{decomposed}||_2$ in a coordinate descent manner. The main idea in the decomposition is that it utilizes the residual $\Wa_{residual} = \Wa - \Wa_{decomposed}$, so that the next iteration approximates the residual tensor $\Wa_{residual}$ by minimizing $||\Wa_{residual} - \Wa_{decomposed}||_2$. This continues until the number of rank-1 tensors found is equal to $R$. More details can be found in \cite{allen2012sparse}.

\subsection{Fine-Tuning} \label{subsec:FineTuning}
As the accuracy will usually drop after decomposition because of the error in decomposition, fine-tuning is needed to recover the accuracy drop. However, as Lebedev et al. \cite{lebedev2014speeding} pointed out, CP decomposition has not yet successfully applied to the whole convolution layers of a CNN with one-time fine-tuning because of its instability \cite{kim2015compression,lebedev2014speeding}.

To overcome the instability, we iteratively fine-tune the whole network after decomposing each layer in order to prevent the errors from getting too big to recover. In the iterative fine-tuning, no layer is frozen because freezing some layers makes the approach greedy, which usually tends to stuck in local minima. As experimented in \cite{yosinski2014transferable}, letting the layers unfrozen shows better results compared to freezing. In this way, all the layers including the already decomposed can adjust to the newly decomposed layer.
\section{Experiments} \label{sec:Experiments}

\begin{table*}[htbp]
	\centering
	\caption{Settings and results comparison with original network and Tucker method of Kim et al. \cite{kim2015compression}. LR: learning rate. Stepsize: frequency (in number of epoch) of decreasing learning rate by factor of 10. Oth.: other layers. 1-2: Conv1 and Conv2.}
	\small
	\begin{tabular}{|c||c|c|c|c|c|c|c|c|r|r|r|r|}
		\hline
		& \multicolumn{2}{c|}{Rank} & \multirow{2}[3]{*}{Epoch} & \multicolumn{2}{c|}{Batch} & \multicolumn{2}{c|}{Learning rate} & \multirow{2}[3]{*}{Stepsize} & \multicolumn{1}{c|}{Top-1} & \multicolumn{1}{c|}{Top-5} & \multicolumn{1}{c|}{\multirow{2}[3]{*}{Weights}} & \multicolumn{1}{c|}{Comp.} \\
		\cline{2-3}\cline{5-8}          & Conv  & FC    &       & 1-2   & Oth.  & 1-2   & Oth.  &       & \multicolumn{1}{c|}{Acc} & \multicolumn{1}{c|}{Acc} &       & \multicolumn{1}{c|}{Cost} \\
		\hline \hline
		\multicolumn{1}{|l||}{Original} & \multicolumn{2}{c|}{-} & \multicolumn{1}{c|}{-} & -     & -     & -     & -     & -     & 56.83  & 79.95  & 61.0M & 724M \\
		\hline
		\multirow{2}[2]{*}{Tucker\cite{kim2015compression}} & \multicolumn{2}{c|}{\multirow{2}[2]{*}{VBMF}} & 15    & \multicolumn{2}{c|}{\multirow{2}[2]{*}{128}} & \multicolumn{2}{c|}{\multirow{2}[2]{*}{0.001}} & \multirow{2}[2]{*}{5} & \multicolumn{1}{c|}{no} & 78.33  & 11.2M & 272M \\
		& \multicolumn{2}{c|}{} & (one-shot) & \multicolumn{2}{c|}{} & \multicolumn{2}{c|}{} &       & \multicolumn{1}{c|}{data} & (-1.62) & ($\times$5.46) & ($\times$2.67) \\
		\hline
		\multirow{2}[2]{*}{This work} & \multirow{2}[2]{*}{150} & \multirow{2}[2]{*}{300} & 15    & \multirow{2}[2]{*}{128} & \multirow{2}[2]{*}{32} & \multirow{2}[2]{*}{0.001} & \multirow{2}[2]{*}{0.002} & \multirow{2}[2]{*}{5} & 54.98  & 78.53  & 8.7M  & 205M \\
		&       &       & (each layer) &       &       &       &       &       & (-1.84) & (-1.42) & ($\times$6.98) & ($\times$3.53) \\
		\hline
	\end{tabular}%
	\label{table:Results}%
\end{table*}%

In this section, we test our approach on AlexNet \cite{krizhevsky2012imagenet}, one of the representative CNNs using Caffe framework \cite{jia2014caffe}. Before describing the main experiments to all of layers, we first briefly introduce AlexNet.
 
\subsection{AlexNet Overview} \label{subsec:AlexnetOverview}
AlexNet is one of famous object recognition architectures and its pre-trained model is available online in Caffe model zoo \cite{jia2014caffe}. As a baseline, we evaluated the accuracy of the pre-trained model using 50,000 validation images from the ImageNet2012 \cite{ILSVRC15} dataset for 1,000 class classification. Top-1 accuracy is 56.83\% and top-5 accuracy is 79.95\%. AlexNet has eight layers in total consisting of five convolution layers and three fully connected layers.  

\subsection{Whole Network Decomposition} \label{subsec:WholeNetworkDecomposition}
In this section, we explain the results of decomposing the whole network with CP-TPM and iterative fine tuning. As mentioned before, to the best of our knowledge, CP-based decomposition has not yet been successful to the whole convolution layer decomposition because of its instability causing the network to be ill-trained. To overcome the problem, we apply fine tuning iteratively so that the errors from each decomposition are not amplified, which usually is with normal one-shot fine tuning performed after the whole convolution layer decomposition.

\begin{figure}[h]
	\centering
	\includegraphics[width=\linewidth]{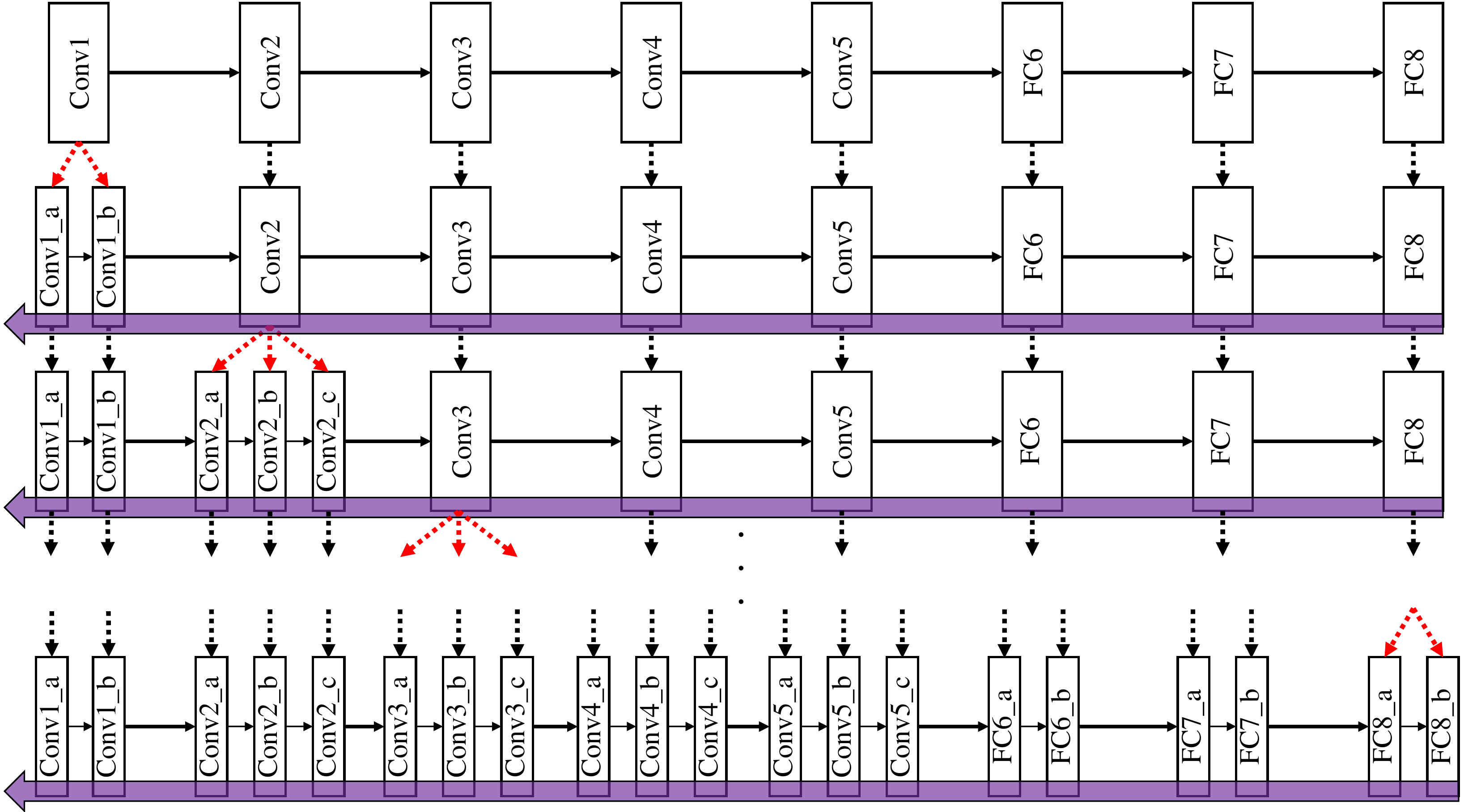}
	\caption{Black solid arrow shows connection between layer. Red dotted line shows the decomposition process while black dotted line means that the weights are taken from the previous iteration. Purple block arrow means fine-tuning by backpropagation to all layers. First, Conv1 is decomposed into 2 layers (red dotted line) while the others remain the same (black dotted line). Then, fine-tuning is performed to all the layers. Afterward, the next layer, Conv2, is decomposed into 3 layers while the others are same. The process repeats until all the layers are decomposed and fine-tuned.}
	\label{fig:GeneralStepAlexnet}
\end{figure}

Fig. \ref{fig:GeneralStepAlexnet} shows the steps of our method in AlexNet. Decomposition and fine tuning are performed for each layer. This process iterates from Conv1 to FC8 in sequence. 

In order to decompose a layer, we first need to figure out the rank. It is desirable for a layer to have as low rank as possible assuming the same level of accuracy. That means each layer should have ranks proportional to its sensitivity, which is defined as the ratio of $loss/total\_loss$.

In order to calculate sensitivity, we first give the same but very small rank (e.g., 5) to all layers and decompose the layers with that rank, which allows us to get the top-5 accuracy loss as in TABLE \ref{tabel:RankOld}. For example, the $total\_loss$ for Conv layers is 57.99, and each convolutional layers is assigned with a rank proportional to its sensitivity value from a total of 750 ranks. The same method is also applied to the fully connected layers and in this case we use a total of 900 ranks, which is higher than for the convolutional layers because the FC layers will certainly have much more effects on accuracy. 

\begin{table}[htbp]
	\centering
	\caption{Rank calculated from the top-5 accuracy loss.}
	\begin{tabular}{|r||l|r|r|r|}
		\hline
		& \multicolumn{1}{c|}{\multirow{2}[1]{*}{Layer}} & \multicolumn{1}{c|}{Top-5} & \multicolumn{1}{c|}{Top-5} & \multicolumn{1}{c|}{\multirow{2}[1]{*}{Rank}} \\
		&       & \multicolumn{1}{c|}{Acc} & \multicolumn{1}{c|}{ Loss} &  \\
		\hline\hline
		
		& Conv1 & 74.57  & 5.38  & 69 \\
		\cline{2-5}          & Conv2 & 68.06  & 11.89  & 154 \\
		\cline{2-5}    \multicolumn{1}{|c||}{Conv} & Conv3 & 68.09  & 11.86  & 153 \\
		\cline{2-5}    \multicolumn{1}{|c||}{(Ave=150)} & Conv4 & 66.27  & 13.68  & 178 \\
		\cline{2-5}          & Conv5 & 64.78  & 15.17  & 196 \\
		\cline{2-5}          & \textbf{Total} &       & \textbf{57.99} & \textbf{750}\\
		\hline
		\hline
		& FC6   & 51.36 & 28.59  & 365 \\
		\cline{2-5}    \multicolumn{1}{|c||}{FC} & FC7   & 58.45 & 21.50  & 275 \\
		\cline{2-5}    \multicolumn{1}{|c||}{(Ave=300)} & FC8   & 59.64 & 20.31  & 260 \\
		\cline{2-5}          & \textbf{Total} &       & \textbf{70.40} & \textbf{900} \\
		\hline
	\end{tabular}%
	\label{tabel:RankOld}%
\end{table}%

As seen in Table \ref{table:Results}, We achieved a better compression results in both the number of weights and computation cost compared with \cite{kim2015compression} while maintaining roughly the same level of accuracy. Specifically, we achieved $1.42\%$ accuracy loss which is less than Tucker accuracy loss of $1.62\%$, with better compression rate. Our method achieved $\times 6.98$ parameter reduction and $\times 3.53$ speeding-up, which is better than Tucker-based method that shows $\times 5.46$ and $\times 2.67$ respectively.

\section{Conclusion} \label{sec:Conclusion}
We have demonstrated that a TPM-based low-rank CP-decomposition combined with the iterative fine-tuning can achieve a whole network decomposition, which has not been tried before with CP. This approach outperforms the previous Tucker-based decomposition method of Kim et al. \cite{kim2015compression} for the whole network decomposition. Specifically, our method achieves $\times 6.98$ parameter reduction and $\times 3.53$ speeding-up in Alexnet, while Tucker-based method shows $\times 5.46$ and $\times 2.67$ respectively.

There remains much work to be further researched. Firstly, there can be many variations in the iterative fine-tuning, for example, freezing layers after fine-tuning and unfreezing in the final fine-tuning; or starting fine-tuning from the last layer. Secondly, this work has focused more on convolution layers than fully connected layers, where we just have applied SVD. Thus, finding the best algorithms for fully connected layers still needs to be tried. Finally and the most importantly, figuring out the ranks in a systematic manner rather than arbitrary will be another key issue in the further research.

\section*{Acknowledgement}
This work was supported by the ICT R\&D program of MSIP/IITP. [B0101-15-0551, Technology Development of Virtual Creatures with Digital Emotional DNA of Users].

{\small
\bibliographystyle{ieee}
\bibliography{egbib}
}

\end{document}